\pdfoutput=1
\documentclass{article}

\usepackage{microtype}
\usepackage{subfigure}
\usepackage{booktabs} 

\usepackage{multirow}
\usepackage[shortlabels]{enumitem}
\usepackage{graphicx}
\usepackage{amsthm}
\usepackage{amsmath}
\usepackage{soul}
\usepackage{algorithm2e}

\newcommand{\fig}[1]{Figure~\ref{fig:#1}}
\newcommand{\sect}[1]{Section~\ref{sect:#1}}
\newcommand{\tab}[1]{Table~\ref{tab:#1}}

\usepackage{hyperref}
\usepackage{amssymb}
\usepackage{pifont}
\usepackage[table]{xcolor}
\usepackage{enumitem}
\usepackage[toc,page]{appendix}

\usepackage[accepted]{innf2021}

\icmltitlerunning{Distilling the Knowledge from Conditional Normalizing Flows}

\begin{document}

\twocolumn[
\icmltitle{Distilling the Knowledge from Conditional Normalizing Flows}




\begin{icmlauthorlist}
\icmlauthor{Dmitry Baranchuk}{to,goo}
\icmlauthor{Vladimir Aliev}{to}
\icmlauthor{Artem Babenko}{to,goo}
\end{icmlauthorlist}

\icmlaffiliation{to}{Yandex, Moscow, Russia}
\icmlaffiliation{goo}{National Research University Higher School of Economics, Russia}

\icmlcorrespondingauthor{Dmitry Baranchuk}{dmitry.baranchuk@graphics.cs.msu.ru}

\icmlkeywords{Machine Learning, ICML}

\vskip 0.3in
]



\printAffiliationsAndNotice{}  

\begin{abstract}
Normalizing flows are a powerful class of generative models demonstrating strong performance in several speech and vision problems. In contrast to other generative models, normalizing flows are latent variable models with tractable likelihoods and allow for stable training. However, they have to be carefully designed to represent invertible functions with efficient Jacobian determinant calculation. In practice, these requirements lead to overparameterized and sophisticated architectures that are inferior to alternative feed-forward models in terms of inference time and memory consumption. In this work, we investigate whether one can distill flow-based models into more efficient alternatives. We provide a positive answer to this question by proposing a simple distillation approach and demonstrating its effectiveness on state-of-the-art conditional flow-based models for image super-resolution and speech synthesis.
\end{abstract}

\section{Introduction}\label{sect:intro}

Normalizing flows (NF)~\cite{Dinh2015NICENI, rezende15}
are a class of generative models that construct complex probability distributions by applying a series of invertible transformations to a simple base density (typically multivariate Gaussian). Such a design allows for exact likelihood computation via change-of-variables formula. Therefore, NF can be straightforwardly trained via likelihood maximization. This NF property is appealing for practitioners since alternative GAN-based models require adversarial optimization, which suffers from vanishing gradients, mode collapse, oscillating or cyclic behavior~\cite{goodfellow2016nips}. 

To enable exact likelihood computation, NF architectures must be composed of invertible modules that also support the efficient calculation of their Jacobian determinant. A large number of such modules have been recently developed, including autoregressive, bipartite, linear and residual transformations~\cite{naf,iaf,maf, sylvester, splines, realnvp, Dinh2015NICENI, rezende15, glow, residualflows}. While some modules are significantly more efficient than others, normalizing flows are generally inferior to feed-forward counterparts (e.g., GANs) in terms of sampling time. In particular, autoregressive flows~\cite{naf, maf} use a slow sequential generation procedure, while bipartite flows~\cite{realnvp, glow} can require a lot of submodules with low expressive power. Moreover, invertibility limits the size of the inner representation leading to impractically deep models. The more comprehensive discussion of the background and related work is deferred to Appendix \ref{appendix:related}.

However, in many applications, one does not need explicit density estimation but requires efficient inference at deployment. This raises a question whether one can relinquish the invertibility of normalizing flows to improve their runtime and memory consumption after training. In this work, we investigate whether this can be achieved through the knowledge distillation approach. In particular, we describe how to distill knowledge from pretrained flow-based models into efficient architectures, which do not suffer from NF design limitations. Our code and models are available online.\footnote{{\footnotesize\url{https://github.com/yandex-research/distill-nf}}}



We summarize our contributions as follows:
\vspace{-2mm}
\begin{itemize}[leftmargin=15px]
    \item We propose a plain training strategy and student design that allow for knowledge distillation from conditional normalizing flows to feed-forward models with streamlined inference and lower memory consumption. To the best of our knowledge, this is the first work that proposes the NF distillation to feed-forward architectures. 
    \item In our experiments, we empirically confirm the effectiveness of our method on the state-of-the-art flow-based models for super-resolution (SRFlow~\cite{srflow}) and speech synthesis (WaveGlow~\cite{waveglow}). We can achieve up to ${\times}10$ speedups with no perceptible loss in quality.
\end{itemize}
\begin{figure*}
\vspace{-3mm}
    \centering
    \includegraphics[width=13.0cm]{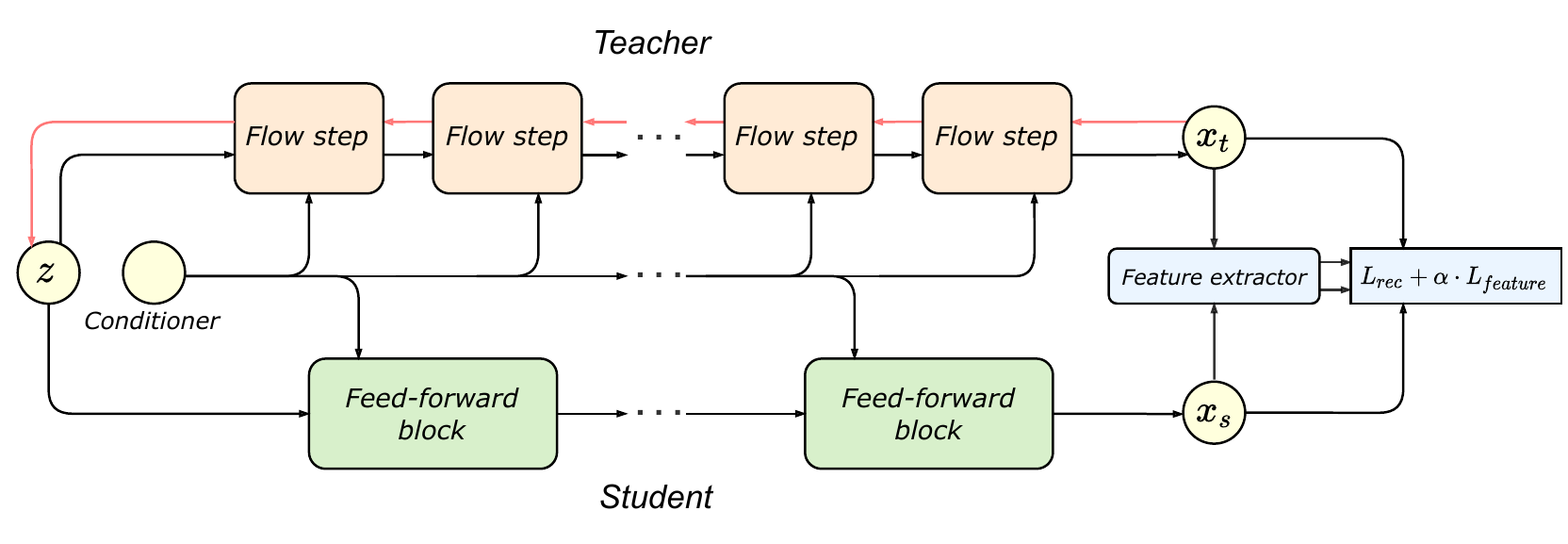}
    \vspace{-5mm}
    \caption{Overall scheme of the proposed knowledge distillation approach. Knowledge is transferred from a flow-based teacher to a feed-forward student using reconstruction and feature losses. As a result, a student is no longer restricted to be a flow and hence can exploit more efficient architectures.}
    \label{fig:scheme}
\vspace{-1mm}
\end{figure*}

\section{Method}\label{sect:method}

This section formally describes the details of our distillation approach. We also provide a general guide how one can design a student architecture for a given flow-based teacher. As illustrative examples, we use two state-of-the-art models from speech and vision domains. 

\subsection{Distillation procedure}\label{sect:procedure}

A pretrained conditional flow-based model defines a deterministic bijective mapping from a noise sample $z{\sim}N(0,{\sigma}I)$ and contextual information $c$ to an output sample $x$. Our distillation approach approximates this mapping by a more efficient model in a supervised manner. The general scheme of our approach is presented in \fig{scheme}.

\subsection{Objective}

The proposed distillation objective is a combination of \textit{reconstruction} and \textit{feature} losses:
\vspace{-1mm}
\begin{equation}
    L = L_{rec} + \alpha \cdot L_{feature} 
\end{equation}
where $\alpha$ is a hyperparameter to balance the loss terms. 

For both applications, the reconstruction loss is an average $L_1$-norm between the student and teacher samples:
\vspace{-1mm}
\begin{equation}
L_{rec}{=}\left\|T_\theta(z, c) - S_\psi(z, c)\right\|_1
\end{equation}
where $T_\theta$ and $S_\psi$ correspond to the teacher and student models, respectively. 

The feature loss for SR model distillation is a perceptual distance between generated images computed by LPIPS~\cite{lpips}. A pretrained VGG16 model~\cite{vgg} is used for LPIPS calculation. 


The feature loss for speech vocoder distillation is a \textit{multi-resolution $STFT$\footnote{$STFT$ stands for the Short-Term Fourier Transform} loss}~\cite{parallel_wavenet, parallel_wavegan,  pddgan}, which is a sum of $STFT$ losses with different parameters (i.e., FFT size, window size and frame shift). In more detail, a single $STFT$ loss is the sum of the \textit{spectral convergence} and \textit{log $STFT$ magnitude} terms~\cite{parallel_wavegan}. 



\begin{figure*}
    \centering
    \includegraphics[width=14.1cm]{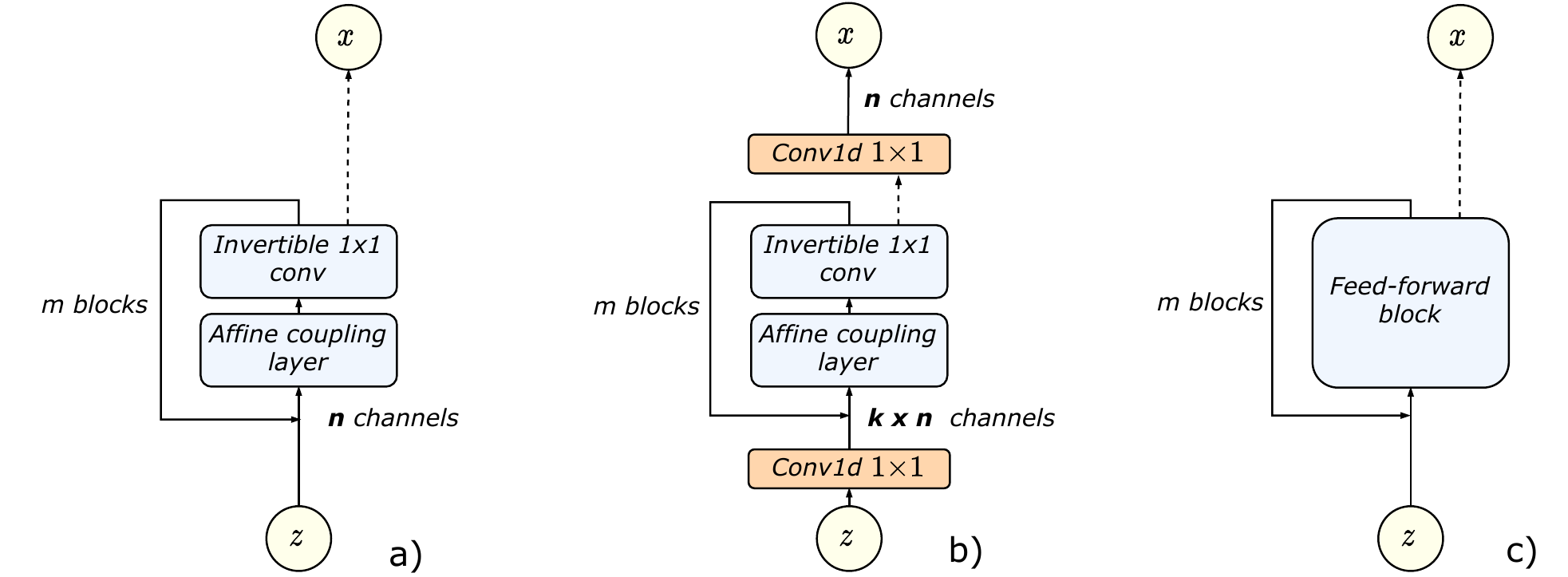}
    \vspace{-1mm}
    \caption{Student design options on the example of NF models with Glow-like flow steps. a) Original flow-based model; b) the same but with increased inner representation dimensions; c) the flow steps are substituted by feed-forward modules.} 
    \label{fig:student_design}
\end{figure*}

\subsection{Student design}
\label{sect:student_design}

The proposed student design can be described by a general recipe: ``just clone a teacher architecture and take advantage of not being a flow anymore''. In other words, our student models inherit the reduced teacher architecture and get free of the following constrains: 

\textbf{Invertibility}. In flow-based models, the size of the inner representations cannot exceed the dimensions of input data vectors. This restriction might lead to much deeper and hence inefficient models. Since the student no longer has to be reversible, one can vary representation dimensions arbitrarily. As a result, student architectures with much fewer blocks can achieve similar performance.

\textbf{Tractable Jacobian determinants}. Flow-based models usually have to exploit specific operations for easy-to-compute Jacobian determinants. Therefore, the student models might benefit from replacing these operations with more efficient and expressive ones. 

Below, we describe the particular design options on the example of flow-based models with Glow-like steps of flow~\cite{realnvp, glow}, which are fundamental building blocks in the SRFlow and WaveGlow models. 


\vspace{-2mm}
\begin{enumerate}[a), leftmargin=15px]
    \item The initial flow-based model, e.g., SRFlow or WaveGlow with fewer parameters.
    
   \item One can consider adjusting the inner representation dimensions using non-invertible $Conv1{\times}1$ layers to increase the overall expressive power.
   
    
    \item Moving on, one can substitute the entire flow step with a commonly used feed-forward module for the corresponding task. Note that feed-forward modules can also vary inner representation dimensions arbitrarily. 
\end{enumerate}

\vspace{-2mm}

In our SRFlow and WaveGlow students, we tune the inner representation size and replace the flow steps with stacked RRDBs~\cite{esrgan} and WaveNet blocks~\cite{wavenet}, respectively. The detailed descriptions of the student architectures are provided in Appendix \ref{appendix:student_arch}. The student design ablation on the example of WaveGlow is presented in Appendix \ref{appendix:ablation}.

\begin{table*}[!t]
\centering
\vspace{-1mm}
\resizebox{0.82\textwidth}{!}{
\begin{tabular}{|c|c|c|c|c|c|c|c|}
     \hline
     Models $\times$4 & PSNR $\uparrow$ & SSIM $\uparrow$ & LPIPS $\downarrow$ & LR-PSNR $\uparrow$ & Param. (M) & Time (ms) & Diversity$\uparrow$\\
     \hline
     RRDB & 29.43 & 0.84 & 0.253 & 49.04 & 16.70 & 303 $\pm$ 1 &  --- \\ 
     \hline
     RankSRGAN & 26.55 & 0.75 & 0.132 & 42.35 & 1.554 & 40 $\pm$ 1 & ---\\ 
     \hline
     ESRGAN & 26.63 & 0.76 & 0.115 & 42.39 & 16.70 & 303 $\pm$ 1 & ---\\ 
     \hline
     \textbf{SRFlow} & \textbf{27.07} & \textbf{0.76} & \textbf{0.120} & \textbf{49.75} & \textbf{39.54} & \textbf{866 $\pm$ 2} & \textbf{0.068} \\ 
     \hline
     \textbf{SRFlow Student} & \textbf{27.32} & \textbf{0.77} & \textbf{0.120} & \textbf{49.52} & \textbf{20.39} & \textbf{420 $\pm$ 1} & \textbf{0.064} \\
     \hline
     \hline
     Models $\times$8 &  PSNR $\uparrow$ & SSIM $\uparrow$ & LPIPS $\downarrow$ & LR-PSNR $\uparrow$& Param. (M) & Time (ms) & Diversity$\uparrow$\\
     \hline
     RRDB & 25.54 & 0.70 & 0.418 & 44.98 & 16.74 & 110 $\pm$ 2 & --- \\
     \hline
     RankSRGAN & --- & --- & --- & --- & --- & --- & ---\\ 
     \hline
     ESRGAN$^*$ & 22.18 & 0.58 & 0.277 & 31.35 & 16.74 & 110 $\pm$ 2 & --- \\ 
     \hline
     \textbf{SRFlow} & \textbf{23.01} & \textbf{0.57} & \textbf{0.271} & \textbf{50.20} & \textbf{50.84} & \textbf{690 $\pm$ 5} & \textbf{0.168} \\
     \hline
    \textbf{SRFlow Student} & \textbf{23.30} & \textbf{0.58} & \textbf{0.269} & \textbf{48.36} & \textbf{16.16} & \textbf{140 $\pm$ 2} & \textbf{0.159} \\
    \hline 
\end{tabular}}

\vspace{-2mm}
\caption{Evaluation metrics on the DIV2K dataset. LPIPS is considered as a primary measure of perceptual quality. The student models provide similar metrics being by ${\times}2.1$ and ${\times}4.9$ faster for ${\times}4$ and ${\times}8$ scaling factors, respectively. (*) denotes that metrics are taken from~\cite{srflow}.}
\label{tab:sr_tab}
\vspace{-2mm}
\end{table*}

\section{Experiments}\label{sect:experiments}

In this section, we compare the learned student against the corresponding teachers and other relevant baselines from the literature. 
 
\subsection{Super-resolution}
\label{sect:exp_sr}
For super-resolution, we compare the student to the following set of models:  

\vspace{-2mm}
\begin{itemize}[leftmargin=15px]
    \item \textbf{ESRGAN}\cite{esrgan}, \textbf{RankSRGAN} \cite{ranksrgan} --- the recent GAN-based super-resolution models. ESRGAN is trained by a combination of reconstruction, perceptual and adversarial losses. RankSRGAN uses the additional ranker network to optimize non-differentiable perceptual metrics.
    
    \item \textbf{RRDB} \cite{esrgan} --- the same as ESRGAN but trained only with $L_1$ objective.
    
    \item \textbf{SRFlow} \cite{srflow} --- a flow-based teacher model. The model is optimized for efficient inference by precomputing inversions for $1{\times}1$ invertible convolutions and fusing actnorm layers.
    
    \item \textbf{SRFlow Student} --- a proposed student model designed according to \sect{student_design}.
\end{itemize}
\vspace{-2mm}

\textbf{Datasets}. The evaluation is performed on the DIV2K dataset~\cite{div2k_dataset} --- one of the established benchmarks for single image super-resolution. In addition to the DIV2K train set, the Flickr2K dataset~\cite{flickr2k_dataset} is also used for training. We consider ${\times}4$ and ${\times}8$ scaling factors between LR and HR images. 

\textbf{Evaluation metrics}. As the primary measure of perceptual performance, we report the established LPIPS~\cite{lpips}, which is proven to correlate with human estimation~\cite{Lugmayr_2020_CVPR_Workshops}. We also report the standard fidelity-oriented metrics, Peak Signal to Noise Ratio (PSNR) and structural similarity index (SSIM)~\cite{ssim}. In addition, we evaluate consistency with the LR image by reporting the LR-PSNR, computed as PSNR between the downsampled SR image and the original LR image.

To evaluate diversity, we follow~\cite{diversity} and measure the average pairwise LPIPS distances between the generated samples for a particular LR image. Larger pairwise distances indicate higher diversity.


\textbf{Evaluation results.} We report metrics as well as corresponding inference time and a number of parameters in \tab{sr_tab}. For most baseline models, we reproduce the results on publicly released checkpoints. Runtimes are measured on a single Tesla V100 on a single validation sample in half precision. The SRFlow student achieves the same metric values as the teacher being by ${\times}2.1$ and ${\times}4.9$ faster for ${\times}4$ and ${\times}8$ scaling factors, respectively. 

Also, we provide qualitative results for ${\times}4$ and ${\times}8$ scaling factors in the appendix, see \fig{div2k_x4} and \fig{div2k_x8}, respectively. The inference for the teacher and student models is performed for the same input noise vectors $z$. We observe that the student produces samples with quality similar to the teacher.

Note that the student preserves the teacher diversity, which indicates that our distillation procedure is potentially able to produce student models with exploitable latent space~\cite{srflow}.

\subsection{Speech Synthesis}\label{sect:exp_speech}

Here, we consider the following set of models for speech synthesis evaluation:

\vspace{-2mm}
\begin{itemize}[leftmargin=15px]
    \item \textbf{WaveGlow}~\cite{waveglow} --- a flow-based teacher model. For efficient inference, we remove  weight norms~\cite{weightnorm} and precompute inversions for convolutional layers. 
    
    
    \item \textbf{NanoFlow}~\cite{nanoflow} --- a recent flow-based vocoder, which produces samples with comparable quality to WaveGlow, but has about ${\times}30$ fewer parameters. 
    
    \item \textbf{MelGAN}~\cite{melgan}, \textbf{Parallel WaveGAN}~\cite{parallel_wavegan}, \textbf{HiFi-GAN}~\cite{hifigan} --- recent GAN-based vocoders. While MelGAN and ParallelWaveGAN are inferior to WaveGlow, HiFi-GAN represents the current state-of-the-art. 
    
    \item \textbf{WG Student} --- a student model designed according to \sect{student_design}. For evaluation, we consider three configurations with $4$/$4$/$2$ WaveNet blocks of $128$/$96$/$96$  hidden channels and denote them as V1/V2/V3, respectively.
\end{itemize}
\vspace{-2mm}

\textbf{Datasets}. All experiments are performed on the LJ speech dataset~\cite{ljspeech}, which is one of the most common benchmarks in speech synthesis. We use a sampling rate of $22,050$kHz and produce mel-spectrograms of the original audio according to~\cite{waveglow}. While \citet{waveglow} originally provide train/validation/test splits for the LJ speech dataset, other methods are not always consistent with them. Therefore, for a fair comparison of the pretrained models, we collect a novel evaluation set and provide its details in Appendix \ref{appendix:testset}.


 
\textbf{Evaluation results.} We compare vocoders in the setting where models are conditioned on ground-truth mel-spectrograms as in previous works~\cite{waveglow, flowavenet}. For all baseline models, we use officially released pretrained models. As a primary metric for generated audio evaluation, we report Mean Opinion Score (MOS) with corresponding $95$ confidence intervals (CI) in \tab{speech}. In Appendix  \ref{appendix:mos}, we describe the detailed protocol used for MOS evaluation. The inference speed is reported in MHz, which stands for $10^6$ audio samples per second. The runtimes are measured on a single Tesla V100 in half precision with sufficiently large batch sizes and sequence lengths to suppress non-model related overheads.

\begin{table}[!t]
\resizebox{0.48\textwidth}{!}{
\begin{tabular}{|c|c|c|c|}
     \hline
     Models & MOS & Param. (M) & Speed (MHz) \\
     \hline
     Ground-truth & 4.50 $\pm$ 0.06 & --- & --- \\
     \hline
     \hline
     NanoFlow & 3.67 $\pm$ 0.09 & 2.82 &  0.369 \\
     \hline
     \textbf{WaveGlow} & \textbf{3.92 $\pm$ 0.08} & \textbf{87.73} & \textbf{1.26}
     \\
     \hline
     \hline
     WG Student V1 & 3.91 $\pm$ 0.08 & 9.53 & 9.36 \\
     \hline
    \textbf{WG Student V2} & \textbf{3.89 $\pm$ 0.09} & \textbf{6.35} & \textbf{12.21}  \\
     \hline
     WG Student V3 & 3.62 $\pm$ 0.09 & 3.18 & 23.62 \\
     \hline
     \hline
     Parallel WaveGAN & 3.84 $\pm$ 0.09 & 1.44 & 2.76  \\
     \hline
     MelGAN & 3.58 $\pm$ 0.08 & 4.27 & 28.40  \\
     \hline
     \textbf{HiFi-GAN} & \textbf{4.10 $\pm$ 0.08} & \textbf{1.46} & \textbf{54.13}  \\
     \hline
\end{tabular}}
\vspace{-2mm}
\caption{Evaluation results on the LJ Speech dataset. WG Student V1 and V2 demonstrate similar speech quality to the WaveGlow teacher being by ${\times}7.4$ and ${\times}9.7$ faster, respectively.
}
\label{tab:speech}
\vspace{-2mm}
\end{table}

V1 and V2 students demonstrate speech quality similar to the teacher and provide ${\times}7.4$ and ${\times}9.7$ faster speech generation, respectively. Moreover, these students have ${\times}9.2$ and ${\times}13.8$ fewer parameters compared to the teacher. 

\section{Conclusion}
\label{sect:conclusion}

In this work, we address the problem of high computational cost of normalizing flows and explain how one can increase the efficiency by giving up invertibility and tractable Jacobian, which are often not necessary for deployed models. In particular, we describe an effective knowledge distillation method from a flow-based teacher to more lightweight student architectures. We empirically demonstrate that the models distilled from conditional normalizing flows are an appealing combination of simplicity, stability and efficiency, needed for typical production pipelines. 


\bibliography{example_paper}
\bibliographystyle{icml2021}

\appendix
\newpage

\begin{appendices}
\begin{figure*}
\vspace{-1mm}
    \centering
    \begin{tabular}{cc}
    \hspace{5mm}\includegraphics[width=6.1cm]{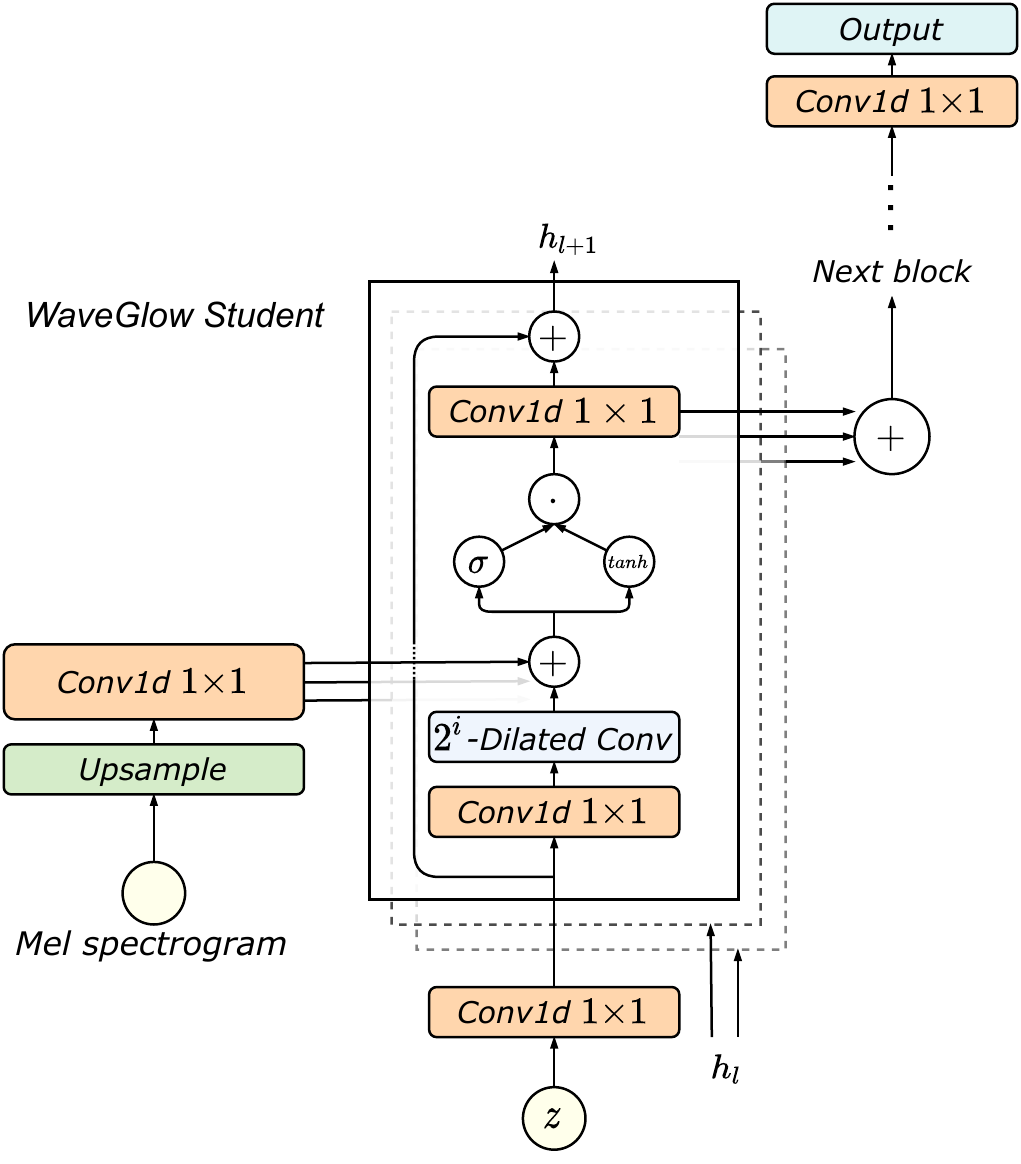} &
    \hspace{20mm}\includegraphics[width=8.1cm, trim=0 -1cm 0 1cm]{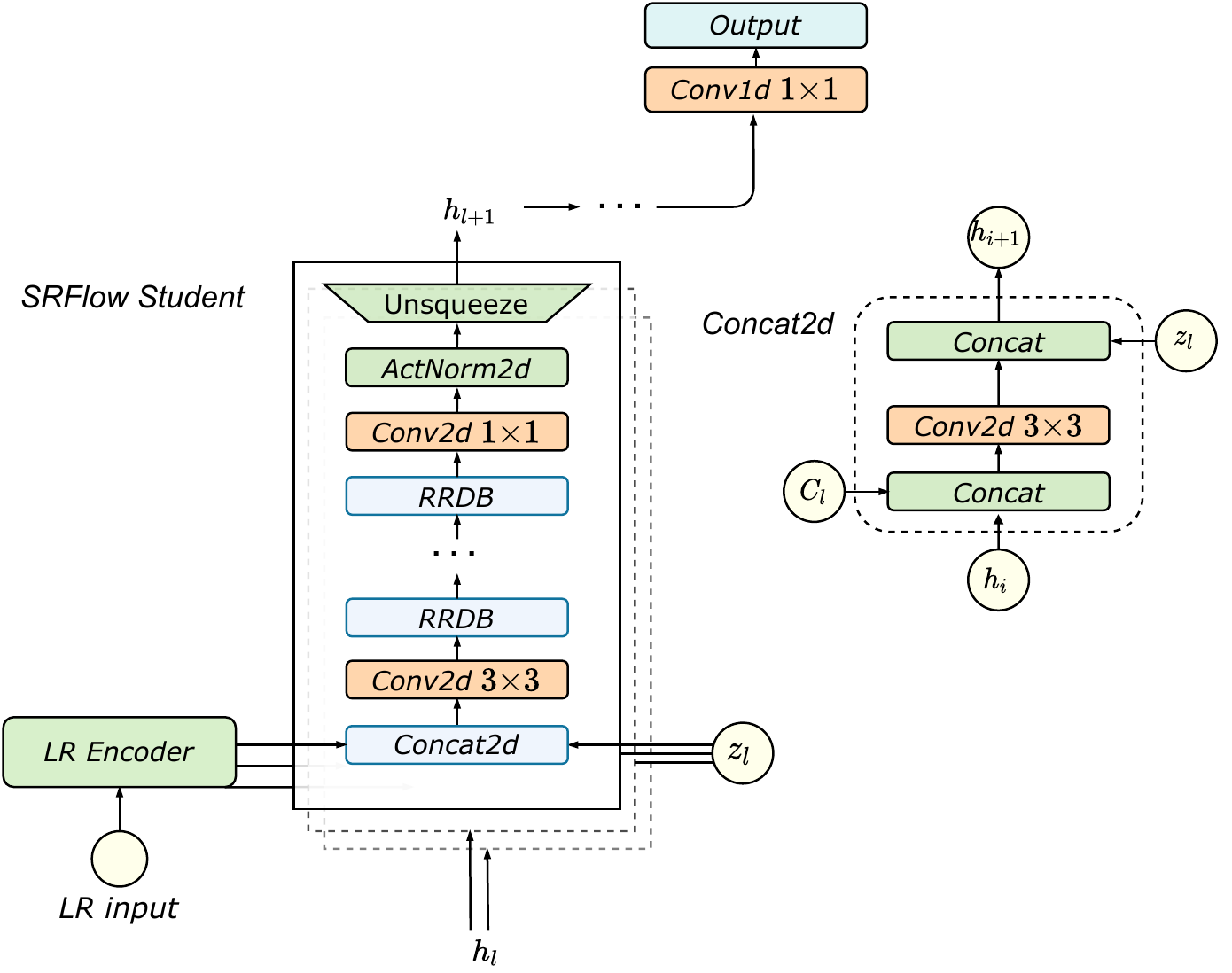}
    \end{tabular}
    
    \vspace{-3mm}
    \caption{\textbf{Left:} WaveGlow student is presented as a sequence of non-causal WaveNet blocks~\cite{wavenet}, which are conditioned on upsampled mel-spectograms. \textbf{Right: }SRFlow student consists of $L$ levels. At each level $l$, feature maps from the previous level are first combined with corresponding LR encoding $c_l$ and noise vector $z_l$. Then, stacked RRDBs followed by an unsqueeze operation are applied.}
    \label{fig:wg_student}
    \vspace{-2mm}
\end{figure*}

\section{Background \& Related work}\label{appendix:related}

This section briefly describes the basics of normalizing flows and their applications in the image super-resolution and speech synthesis tasks. Finally, we provide a short overview of knowledge distillation techniques.

\subsection{Normalizing flows}

Normalizing flows rely on change-of-variable formula to compute exact log likelihood as a sum of Jacobian log-determinants. In general, computing the Jacobian determinant of a transformation with $N$-dimensional inputs and outputs has $O(N^3)$ computational complexity, which is intractable for large $N$ in typical deep learning applications. Therefore, one of the challenges is to design invertible architectures with efficient determinant calculation. 

Currently, there are two main families of flow-based models with easy-to-compute Jacobian determinants: based on autoregressive and bipartite transformations. The autoregressive family includes \textit{autoregressive flow} (AF)~\cite{maf, naf} and \textit{inverse autoregressive flow} (IAF)~\cite{iaf}. AFs resemble autoregressive models~\cite{uria2016neural}, which allow for parallel density estimation but perform sequential synthesis. In contrast, IAF poses parallel synthesis but sequential density estimation, making likelihood-based training very slow. The second family includes \textit{bipartite transformations}~\cite{realnvp, glow}, which provide both efficient likelihood-based training and parallel synthesis. However, bipartite flows are usually less expressive than the autoregressive ones and hence the models require a larger number of layers and parameters. Recent works develop more powerful and efficient normalizing flows. ~\citet{waveflow, nanoflow} combine the ideas of autoregressive and bipartite transformations. Others propose more expressive invertible layers and architectures~\cite{ho2019flow, splines, invconvolutions, residualflows, iresnet} or continuous transformations~\cite{ffjord, neuralode}.


At the moment, conditional NFs are gaining popularity in various practical speech and vision applications. In particular, several recent works~\cite{waveglow, flowavenet, srflow, pointflow} exploit the ideas of Glow~\cite{glow}, RealNVP~\cite{realnvp} and continuous flows~\cite{neuralode} for waveform synthesis, image super-resolution, and point cloud generation. In this work, we focus on the state-of-the-art flow-based models for image super-resolution~\cite{srflow} and speech synthesis~\cite{waveglow}.

\subsection{Super-resolution}

Super-resolution (SR) is one of the fundamental image processing problems which aims to improve the quality of low-resolution (LR) images by upscaling them to high-resolution (HR) ones with natural high-frequency details.

Single image super-resolution approaches tend to either discover the most efficient neural network architectures~\cite{rdn,rcan,scsrmd} or improve high-frequency details generation by introducing more sophisticated objectives~\cite{srgan, esrgan, ranksrgan}. In contrast to these approaches, the SRFlow model~\cite{srflow} designs a flow-based architecture that is trained by likelihood maximization. As a result, SRFlow estimates a conditional distribution of natural HR images corresponding to a given LR image. 

SRFlow is built upon the Glow~\cite{glow} architecture that gradually transforms a sample from initial distribution $z{\sim}N(0, I)$ into the HR image. The model is conditioned on representations of LR images delivered by a deterministic encoder~\cite{esrgan}.



\subsection{Speech Synthesis}

The state-of-the-art performance for speech synthesis is also  achieved by deep generative models. In typical speech synthesis pipelines, neural vocoders\footnote{vocoder --- speech waveform synthesizer} play one of the most important roles. Usually, a neural vocoder synthesizes time-domain waveform and is conditioned on mel-spectrograms from a text-to-spectrogram model~\cite{deepvoice1, deepvoice3, fastspeech, fastspeech2, tacotron2}. 

Long-standing state-of-the-art neural vocoders are autoregressive models~\cite{wavenet, wavernn, samplernn} that provide impressive speech quality but suffer from  slow sequential generation. Therefore, a wide line of works aims to speedup their inference. 

Flow-based models have been successfully applied for parallel waveform synthesis with fidelity comparable to autoregressive models~\cite{parallel_wavenet, waveglow, waveflow, nanoflow, flowavenet, clarinet, blow}. Among flow-based vocoders, WaveGlow~\cite{waveglow}, WaveFlow~\cite{waveflow}, NanoFlow~\cite{nanoflow} and FloWaveNet~\cite{flowavenet} have a particular emphasis on efficiency. In this work, we specifically consider the WaveGlow model~\cite{waveglow} --- one of the current state-of-the-art flow-based vocoders. 


Being efficient enough, GAN-based vocoders~\cite{melgan, parallel_wavegan} used to provide inferior speech fidelity compared to the state-of-the-art flow-based models. However, recently proposed HiFi-GAN~\cite{hifigan} efficiently delivers high-quality speech due to the training procedure with multi-scale~\cite{melgan} and multi-period discriminators~\cite{hifigan}.


An alternative line of works proposes vocoders based on the denoising diffusion framework~\cite{diffwave, wavegrad}. These models demonstrate promising speech quality but are not efficient enough for many usage cases.

\subsection{Knowledge Distillation}

Knowledge distillation is one of the most popular compression and acceleration techniques for large models and ensembles of neural networks~\cite{hinton_kd, noise_kd, romero2015}. Specifically, the idea of knowledge distillation is to train a single efficient student on predictions produced by a computationally expensive teacher. 

The closest works to ours distill knowledge from an expensive autoregressive neural vocoder to normalizing flows with parallel inference. In particular, Parallel WaveNet~\cite{parallel_wavenet} and ClariNet~\cite{clarinet} transfer knowledge from a pretrained WaveNet~\cite{wavenet} to IAF~\cite{iaf}. As opposed to these works, we address the computational inefficiency of flow-based models and propose to transfer knowledge from normalizing flows to feed-forward networks.

\section{Student architectures}\label{appendix:student_arch}
Below, we describe the architectures for WaveGlow and SRFlow students in more detail and depict them in \fig{wg_student} (Left) and (Right), respectively. 

\subsection{SRFlow Student} 

\textbf{LR encoder.} Both teacher and student models condition on the LR representations produced by the LR encoder.  The LR encoder is a feed-forward SR architecture based on $23$ Residual-in-Residual Dense Blocks (RRDB)~\cite{esrgan}. Originally, RRDBs are composed of $3$ Residual Dense Blocks (RDB) and each RDB consists of $5$ convolutional layers. In our student models, we vary the numbers of RDB blocks as well as convolutional layers within them.

The LR encoder is unchanged for ${\times}4$ scaling factor. For ${\times}8$ scaling factor, the RRDB module is substituted by a single RDB with $3$ convolutional layers.

\textbf{Base network.} The base architecture consists of $L$ levels. At each level $l$, the model combines the corresponding LR representation $C_l$, the activations from the previous level $h_l$ and the noise vector $z_l$ by the \texttt{Concat2d} layer. Here, one can vary $h$ channel dimensions, according to \sect{student_design} b).

Then, instead of a sequence of flow steps in the teacher model, stacked RRDB modules follow. 
For ${\times}4$ scaling factor, we use $6$ RRDB modules of $2$ RDBs with $3$ convolutional layers. For ${\times}8$, a single original RRDB is used. 

Finally, similarly to the SRFlow teacher, the transition step and unsqueeze operation are applied. The transition step is Conv$1{\times}1$ followed by the Actnorm2d layer~\cite{glow}. The unsqueeze operation doubles the spatial resolution of feature maps and reduces their channel dimension by a factor of $4$.

\subsection{WaveGlow Student} 

WaveGlow student represents a sequence of conditional non-causal WaveNet blocks~\cite{wavenet}. We start and finish with Conv$1{\times}1$ layers to adjust the input channel dimension. Each WaveNet block is organized into $8$ residual layers, each consisting of a dilated convolution followed by a gated activation unit~\cite{pixelcnn} and $1{\times}1$ convolution. The upsampled mel-spectrograms are added to the intermediate activations before the gated activation unit. In contrast to the teacher, a student does not inject noise between intermediate WaveNet blocks but obtains the entire $z$ at the very beginning.

\section{Training details}

\textbf{SRFlow students.} The models are trained on patches of size $128{\times}128$ by the Adam optimizer~\cite{kingma2014adam} with learning rate of $2e{-}4$ for $100$ epochs. The learning rate is dropped by a factor of $2$ at each $25$-th epoch. The batch sizes of $80$ and $128$ are used for $4$ and $8$ scaling factors, respectively. The loss coefficient $\alpha$ is $10$. 

\textbf{WaveGlow students.} The models are optimized on short segments of length $16000$ samples by the Adam optimizer~\cite{kingma2014adam} with a learning rate of $1e{-}4$ for $200000$ iterations. For faster convergence, we use One Cycle learning rate schedule~\cite{onecycle} with $5000$ warm up steps. The batch size is $256$. The loss coefficient $\alpha$ is $1e{-}2$ for most student settings.

All hyperparameters are tuned on the hold-out train samples, which do not participate in the final evaluation. The training is performed on $8{\times}$ Tesla V100.

\section{Mean Opinion Score}\label{appendix:mos}
We crowd-sourced MOS tests on $400$ English-speaking individuals per model. Each rater had to pass training on $21$ validation audio samples from various models with golden results and corresponding clarifications. Then, they listened to $10$ audio samples from a single model and rated their naturalness on a five-point scale, see \tab{scalemos}. Each rater was asked to wear headphones and work in a quiet environment. To aggregate MOS results for each model, we follow the protocol from~\cite{melgan}.

\begin{table}[!ht]
\resizebox{0.48\textwidth}{!}{
\begin{tabular}{|c|c|}
     \hline
     \textbf{Rating} & \textbf{Quality} \\
     \hline
     \multirow{2}{*}{5} & Natural speech. \\ 
     & Distortions are imperceptible.\\ 
     \hline
     \multirow{2}{*}{4} & Mostly natural speech. \\ 
     & Distortions are slightly perceptible and rare.\\ 
     \hline
     \multirow{2}{*}{3} &  Equally natural and unnatural speech. \\
     & Distortions are perceptible and almost permanent.\\
     \hline
     \multirow{2}{*}{2} &  Mostly unnatural speech. \\
     & Distortions are annoying, but not objectionable.\\
     \hline
     \multirow{2}{*}{1} & Completely unnatural speech. \\
     & Distortions are very objectionable.\\
     \hline
\end{tabular}}
\vspace{-3mm}
\caption{The rating scale presented to $400$ individuals for MOS evaluation. This rating scale was adapted from~\cite{mos} to better fit modern speech synthesis applications. }
\label{tab:scalemos}
\end{table}

\section{Evaluation set for Speech Synthesis}\label{appendix:testset}
The collected set consists of $50$ utterances trimmed from publicly available recordings by Linda Johnson from LibriVox. We carefully choose recordings to be similar to the original dataset and normalize audio samples accordingly. This test set will be publicly available online for reproducibility purposes.

\begin{table}[!t]
 \resizebox{0.48\textwidth}{!}{
\begin{tabular}{|c|c|c|c|}
     \hline
     Models & MOS & Param. (M) & Speed (MHz) \\
     \hline
     Flow Student & 3.42 $\pm$ 0.08 & 6.28 & 12.25  \\
     \hline
     Wide Flow Student & \textbf{3.93 $\pm$ 0.08} & 6.37 & 11.95  \\
     \hline
     WG Student & \textbf{3.89 $\pm$ 0.09} & 6.35 & 12.21 \\
    \hline
\end{tabular}}
\vspace{-3mm}
\caption{The comparison of the student design principles described in \sect{student_design}. Both options provide essentially the same performance. This indicates that relaxing the invertibility constraint is sufficient to achieve significant improvements.}
\label{tab:speech2}
\end{table}

\section{Student design ablation}\label{appendix:ablation}

To identify the core change to the architecture that improves the student performance, we ablate the student design principles described in \sect{student_design} on the speech synthesis task. In this experiment, we consider the student models with $4$ WaveNet blocks of $96$ channels and evaluate the following configurations in correspondence with \fig{student_design}: 

\begin{enumerate}[a), leftmargin=15px]
    \item \textbf{Flow Student} --- a student corresponds to the smaller WaveGlow model with all NF restrictions mentioned in \sect{student_design}.
    
    \item \textbf{Wide Flow Student} --- a flow student where the inner representation channel dimension is increased from $8$ to $96$. Note that this model is not invertible anymore. 
    
    
    \item \textbf{WG Student} --- a student model where flow steps are replaced with WaveNet blocks.
\end{enumerate}

According to \tab{speech2}, once the invertibility constraint is lifted, the corresponding student can achieve significantly better speech quality compared to the flow student of the same capacity.

\begin{figure*}
    \centering
    \includegraphics[width=17.2cm]{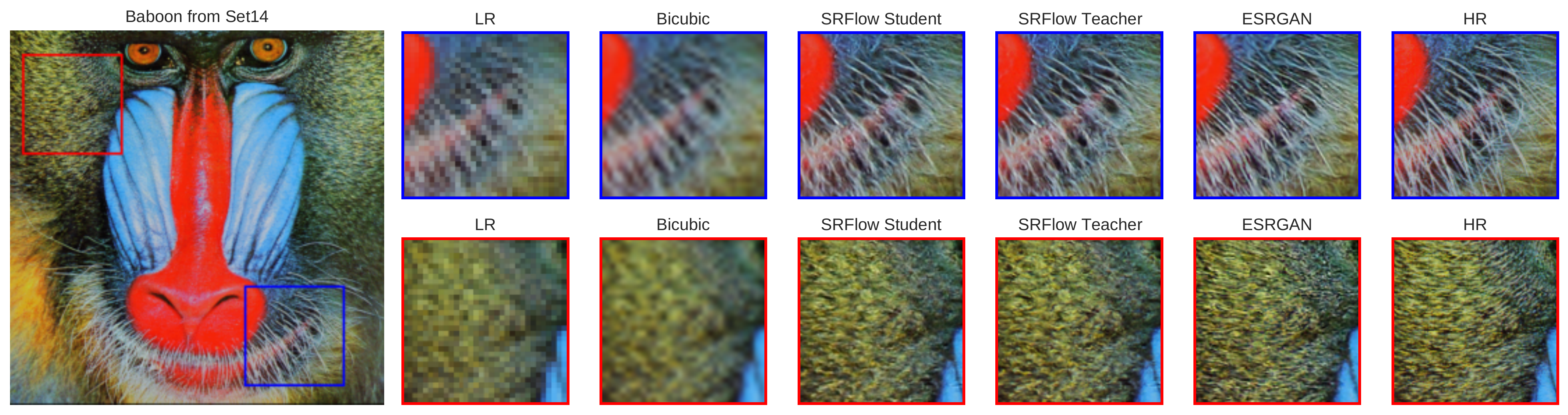}\\
    \includegraphics[width=17.2cm]{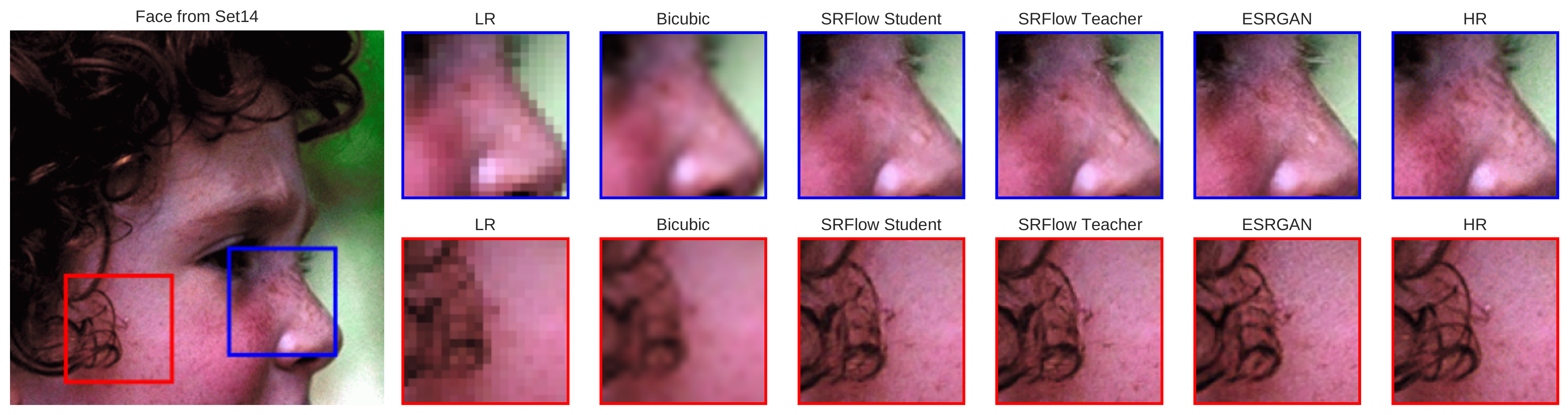}\\
    \includegraphics[width=17.2cm]{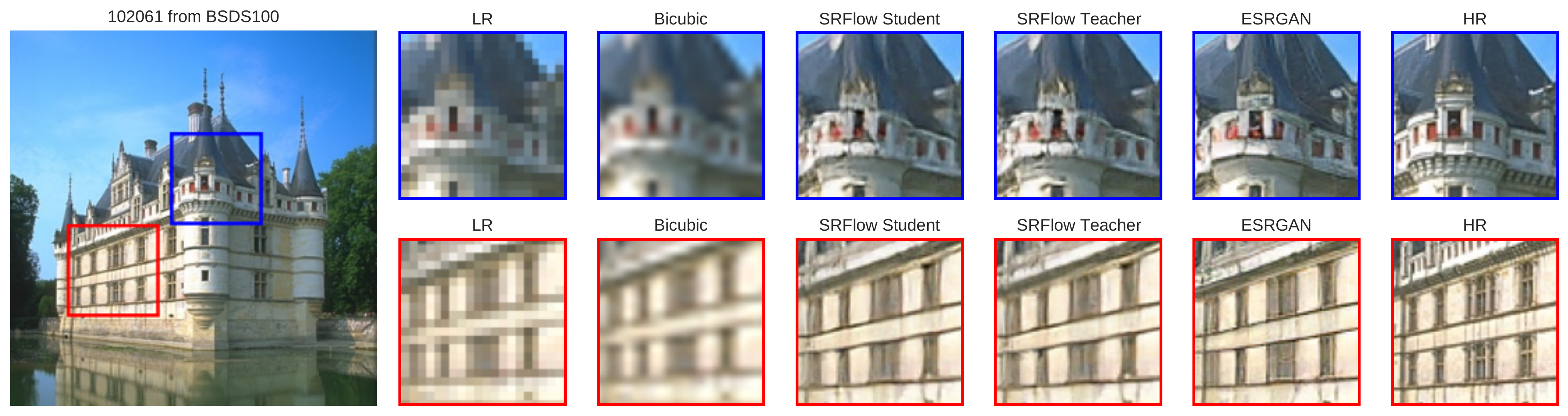}\\
    \includegraphics[width=17.2cm]{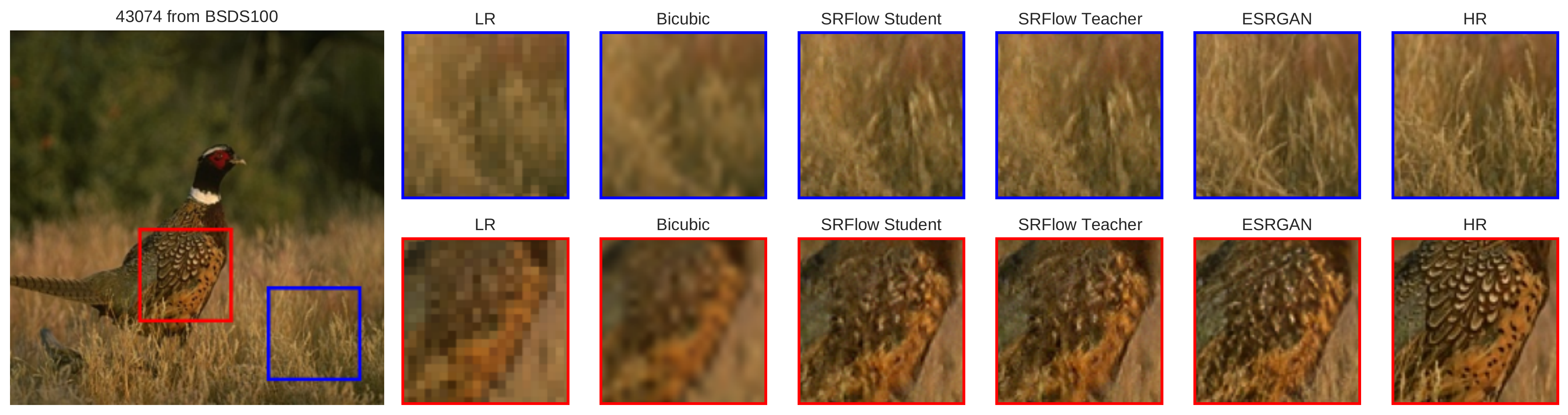}\\

    \caption{Qualitative results of SRFlow teacher and student for ${\times}4$ scaling factor. The student model produces SR samples similar to the teacher ones.}
    \label{fig:div2k_x4}
\end{figure*}

\begin{figure*}
    \centering
    \includegraphics[width=17.2cm]{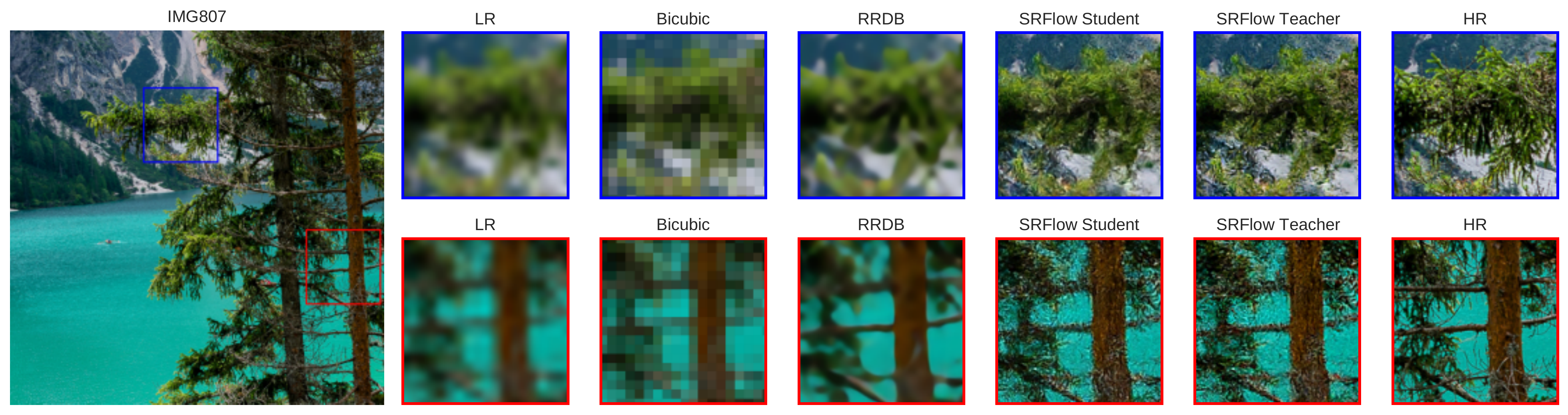}\\
    \includegraphics[width=17.2cm]{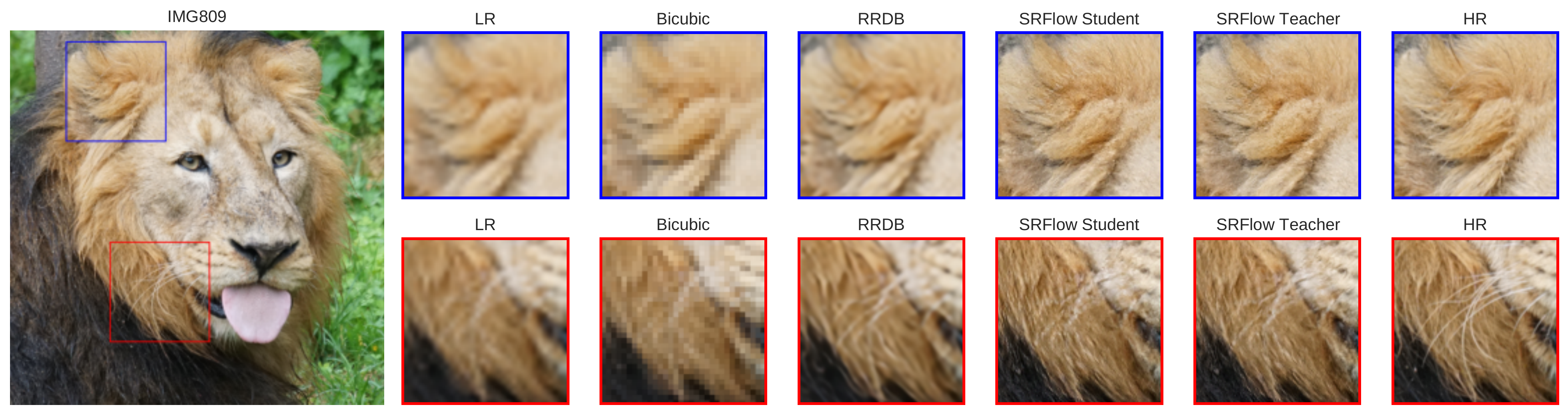}\\
    \includegraphics[width=17.2cm]{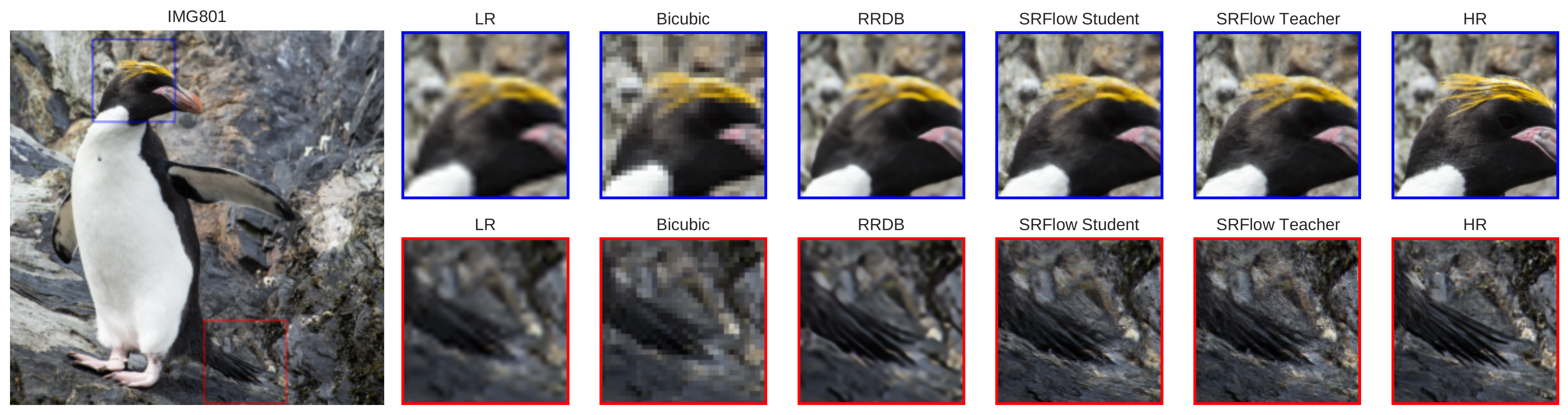}\\
    \includegraphics[width=17.2cm]{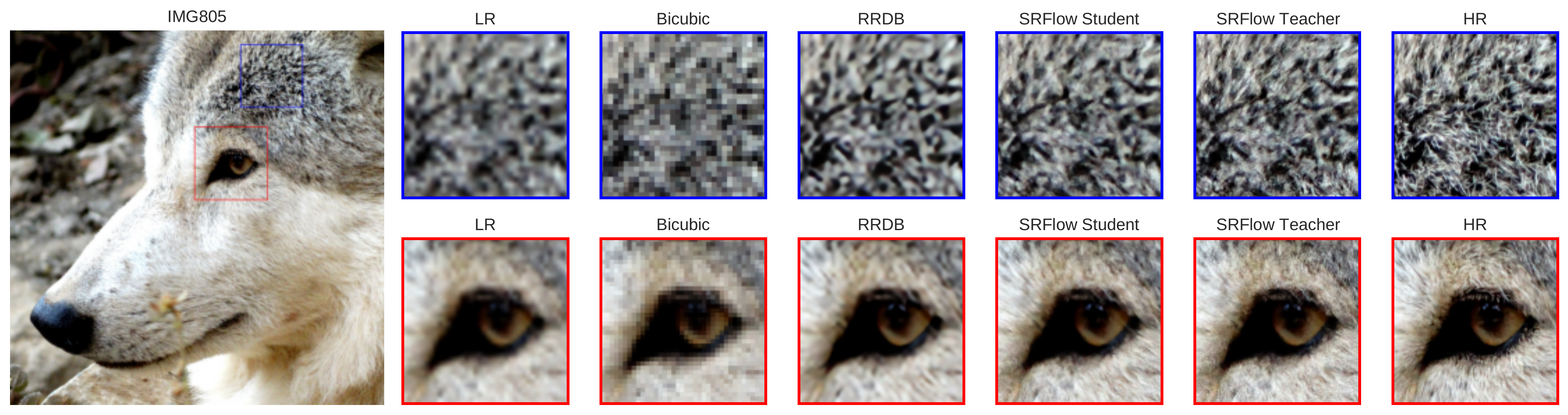}\\

    \caption{Qualitative results of SRFlow teacher and student on the DIV2K dataset for ${\times}8$ scaling factor. On most samples, the student model demonstrates performance close to the teacher.}
    \label{fig:div2k_x8}
\end{figure*}

\end{appendices}

\end{document}